\def\year{2020}\relax
\documentclass[letterpaper]{article} 
\usepackage{aaai20}  
\usepackage{times}  
\usepackage{helvet} 
\usepackage{courier}  
\usepackage[hyphens]{url}  
\usepackage{graphicx} 
\usepackage{graphicx}  
\usepackage{booktabs} 
\usepackage{latexsym}
\usepackage{amssymb}
\usepackage[utf8]{inputenc}
\usepackage{enumitem}
\usepackage{todonotes}
\usepackage{amsmath}
\usepackage{array}
\usepackage{wasysym}
\usepackage{multirow}
\usepackage{algorithm}
\usepackage[noend]{algpseudocode}

\title{Understanding Chat Messages for Sticker Recommendation in Messaging Apps}
\author{Abhishek Laddha, Mohamed Hanoosh, Debdoot Mukherjee, Parth Patwa, Ankur Narang \\ Hike Messenger \\  \{abhishekl, moh.hanoosh, debdoot, parthp, ankur\}@hike.in
}
 
\begin{document}

\maketitle
\begin{abstract}

Stickers are popularly used in messaging apps such as Hike to visually express a nuanced range of thoughts and utterances to convey exaggerated emotions. However, discovering the right sticker from a large and ever expanding pool of stickers while chatting can be cumbersome.
In this paper, we describe a system for recommending stickers in real time as the user is typing based on the context of the conversation. We decompose the sticker recommendation (SR) problem into two steps. First, we predict the message that the user is likely to send in the chat. Second, we substitute the predicted message with an appropriate sticker. 
Majority of Hike's messages are in the form of text which is transliterated from users' native language to the Roman script. This leads to numerous orthographic variations of the same message and makes accurate message prediction challenging. To address this issue, we learn dense representations of chat messages employing character level convolution network in an unsupervised manner. We use them to cluster the messages that have the same meaning. In the subsequent steps, we predict the message cluster instead of the message. 
Our approach does not depend on human labelled data (except for validation), leading to fully automatic updation and tuning pipeline for the underlying models. 
We also propose a novel hybrid message prediction model, which can run with low latency on low-end phones that have severe computational limitations. Our described system has been deployed for more than $6$ months and is being used by millions of users along with hundreds of thousands of expressive stickers. 

\end{abstract}

\section{Introduction}
\label{sec:intro}

\begin{figure}[t!]
    \includegraphics[width=0.95\columnwidth]{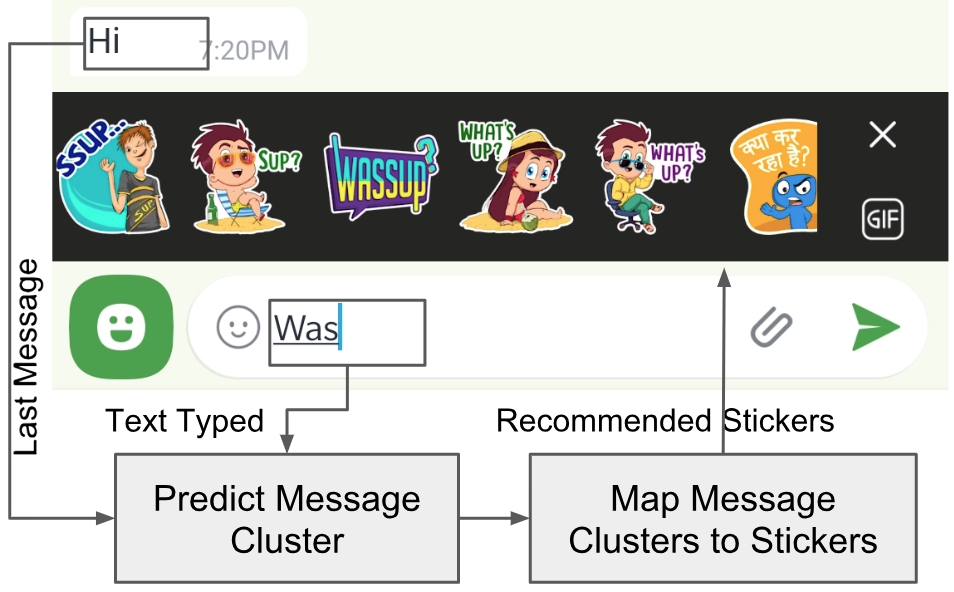}
    \caption{Sticker Recommendation (SR) UI on Hike and a high level flow of our 2 step SR system.}
    \label{fig:problem}
\end{figure}

\frenchspacing
In messaging apps such as Facebook Messenger, WhatsApp, Line and Hike, new modalities are extensively used to visually express thoughts and emotions (\textit{e.g.} emojis, gifs and stickers). 
Emojis ({\em e.g.,} \smiley, \frownie) are used in conjunction with text to convey emotions in a message \cite{donato2017investigating}. Unlike emojis, stickers provide a graphic alternative for text messages. Hike stickers are composed of an artwork (e.g., cartoonized characters and objects) and a stylized text (See Fig. \ref{fig:problem}). They help to convey rich expressions along with the message. Hundreds of thousands of stickers are available for free download or purchase on sticker stores of popular messaging apps. Once a user downloads a sticker pack, it gets added to a palette, which can be accessed from the chat input box. However, discovering the right sticker while chatting can be cumbersome because it's not easy to think of the best sticker that can substitute your utterance.
Apps like Hike and Line offer type-ahead SR while typing (as shown in Fig. \ref{fig:problem}) in order to alleviate this problem. In comparison to emoji prediction which predicts a few set of emotions, there are numerous possible utterances in text (in tens of thousands) and their corresponding stickers which makes SR problem more complex than emoji prediction \cite{barbieri2017emojis}. 

The latency of generating such SR should be in tens of milliseconds in order to avoid any perceivable delay during typing. This is possible only if the system runs end-to-end on the mobile device without any network calls. Furthermore, a large fraction of Hike users use low-end mobile phones, so we need a solution which is efficient both in terms of CPU load and memory requirements. 


\par
Prior to this work, SR on Hike app was based on string matching of the typed text to the tags that were manually assigned to each sticker. 
Recall of string matching based approach is limited by exhaustiveness of tagging. 
However, there are many ways of expressing same message in a chat. For instance, people often skip vowels in words as they type; {\em e.g.,} ``where are you'' $\rightarrow$ ``whr r u'', ``ok'' $\rightarrow$ ``k''. This is further exacerbated when people transliterate messages from their native language to Roman script; {\em e.g.,} phrase ``acchha'' (Hindi for ``good'') is written in many variants - ``accha'', ``acha'', ``achha'' {\em etc.} Another reason for proliferation of such variants is that certain words are pronounced differently in different regions. We observe 343 orthographic variants of ``kya kar raha hai'' (Hindi for ``What are you doing'') in our dataset. Hence, it is hard for any person to capture all variants of an utterance as tags. 
\\
{\em \bf \noindent Decomposing SR:} SR can be set up as a supervised task by directly learning the most relevant stickers for a given context defined by the previous message and the text typed by the user. However, due to frequent updates in the set of available stickers and a massive skew in historical usage towards a handful of popular stickers, it becomes difficult to collect unbiased data to train such an end-to-end model. 
Thus, we decompose the SR task into 2 steps. First, we predict the message that a user is likely to send based on the chat context and what the user has typed. 
Second, we recommend stickers by mapping the predicted message to stickers that can substitute it. 
We automatically update the mapping frequently based on relevance feedback observed on recommendations and incorporate new stickers as they are launched.

In this paper, we focus on how to efficiently set up the {\em message prediction} using classification model. For doing so, we propose following solutions:

{\em \noindent \textbullet \hspace{1pt} Chat Message Clustering:} 
As described above, a large number of chat messages are simply variants of each other. Similar to SmartReply \cite{kannan2016smart} which clusters the short responses having similar intent, we cluster frequent messages which are orthographic variants or semantically similar. But unlike their approach to apply semi-supervised learning for clustering, we learn an embedding of chat message in an unsupervised manner. Then, we perform clustering on the representations with the help of HDBSCAN \cite{mcinnes2017hdbscan} algorithm. We investigate various encoders to learn the embeddings and empirically show that the use of charCNN \cite{kim2016character} with transformer \cite{vaswani2017attention} can be highly effective to capture semantics of chat phrases. The clusters obtained are used as classes for our message prediction model. This helps us in drastically reduce the number of classes in the classifier while keeping most frequent message intent of our corpus.
{\em \noindent \textbullet \hspace{1pt} Hybrid Message prediction model for low-end smart-phones:} 
Running inference with a neural network model (NN) for message prediction proves to be challenging on low-end mobile devices with severe memory limitations. \cite{gysel2016hardware}. 
The size of a NN model trained for message prediction exceeds the memory limitation even after quantization. We present a novel hybrid model, which can run efficiently on low-end devices without significantly compromising accuracy. Our system is a combination of a NN based model, running on the server, that processes chat context predicts message cluster, and a Trie based model that processes typed text input on the client. The first component is not limited by memory and CPU. Trie search is an efficient mechanism to retrieve message cluster based on typed text and
can be executed for each character typed. Hence, the system can satisfy the latency constraints with this setup. Scores from these two components are combined to produce final scores for message prediction.
Lastly, we demonstrate the efficacy of our proposed system on both offline evaluation and real world deployment performance. We discuss other challenges in deployment such as multiple languages in India, serving and update frequency of models. 
In summary, our contributions include:
\begin{itemize}[leftmargin=*]
\item We present 
a novel system for type-ahead SR within a messaging application. Our deployed system automatically updates and tunes itself using updated online chat corpus without needing any manual intervention. We decompose the SR task in 2 steps: {\em message prediction} and {\em sticker substitution}.
\item We describe 
an unsupervised approach to cluster chat messages with similar semantics. We evaluate different encoders to learn semantic representation of messages.
\item We propose 
a novel hybrid message prediction model, which can run efficiently with low latency and low memory footprint on low-end mobile phones. 
We demonstrate that this hybrid model has comparable performance with a server only NN based message prediction model.
\end{itemize}

 
\section{Related Work}

\frenchspacing
To the best of our knowledge there is no prior art which studies the problem of type-ahead SR. However, there are a few closely related research threads that we describe here. 

The use of emojis is widespread in social media. 
\citeauthor{barbieri2018multimodal} \shortcite{barbieri2018multimodal} predict which of the top-20 emojis are likely to be used in an Instagram post based on the post's text and image. Unlike emojis that are majorly used in conjunction with text, stickers are independent messages that substitute text. Thus, in order to have effective SR for a given conversational context, we need to predict the likely utterance and not just the emotion. Since the possible utterances are more numerous than emotions, our problem is more nuanced than emoji prediction.

There exists a large body of research on conversational response generation. 
\citeauthor{xing2017hierarchical} \shortcite{xing2017hierarchical}, \citeauthor{serban2016building} \shortcite{serban2016building} design an end-to-end model leveraging a hierarchical RNN to encode the input utterances and another RNN to decode possible responses. 
\citeauthor{zhang2018generating} \shortcite{zhang2018generating} describe a model that explicitly optimizes for improving diversity of responses. 
\citeauthor{yan2016learning} \shortcite{yan2016learning} proposed a retrieval based approach with the help of a DNN based ranker that combines multiple evidences around queries, contexts, candidate postings and replies.
Smart Reply \cite{kannan2016smart} proposed a system that suggests short replies to e-mails which are high quality as well as diverse. 
Akin to our system, SmartReply also generates clusters of responses with same intent. They apply semi-supervised learning to expand the set of responses starting from a small number of manually labeled responses for each semantic intent. In contrast, we follow an unsupervised approach to discover message clusters since the set of all intents that may correspond to stickers aren't readily available. A unique aspect of our system is that we update the message prediction by incorporating whatever the user has typed so far; we need to do this in order to deliver type-ahead SR.

There is a parallel research thread around learning effective representations for sentences that can capture sentence semantics. Skip Thought \cite{kiros2015skip} is an encoder-decoder model that learns to generate the context sentences for a given sentence. It includes sequential generation of words of the target sentences, which limits the target vocabulary and increases training time. Quick Thought \cite{logeswaran2018efficient} circumvents this problem by replacing the generative objective with a discriminative approximation, where the model attempts to classify the embedding of a correct target sentence given a set of sentence candidates. Recently, \citeauthor{devlin2018bert} \shortcite{devlin2018bert} proposed BERT which predicts the bidirectional context to learn sentence representation using transformer \cite{vaswani2017attention}. \citeauthor{yang2018learning} \shortcite{yang2018learning} proposes the Input-Response model that we have evaluated in this paper. However, unlike these works, we apply a CharCNN \cite{zhang2015character,kim2016character} in our encoder to deal with the problem of learning similar representations for phrases that are orthographic variants of each other.


\section{Chat Message Clustering}
\label{sec:groupformation}

\frenchspacing
As mentioned earlier, we cluster frequent messages in our chat and use them as classes in the message prediction model. For covering a large fraction of messages in our chat corpus with a limited number of clusters, we need to group all messages with the same intent into a single cluster. This has to be done without compromising the semantics of each cluster. In order to efficiently cluster messages, it is critical to obtain dense vector representations of messages that effectively capture their meaning.

In this section, we describe an encoder which is used to convert chat phrases into dense vectors \cite{mikolov2013distributed}. Then, we explain the network architecture which is used to train the message embeddings such that we learn similar representations for messages having same meaning. 
\subsection{Encoder}
\begin{figure}[t!]
    \centering
    \includegraphics[width = 7cm, height=7.5cm]{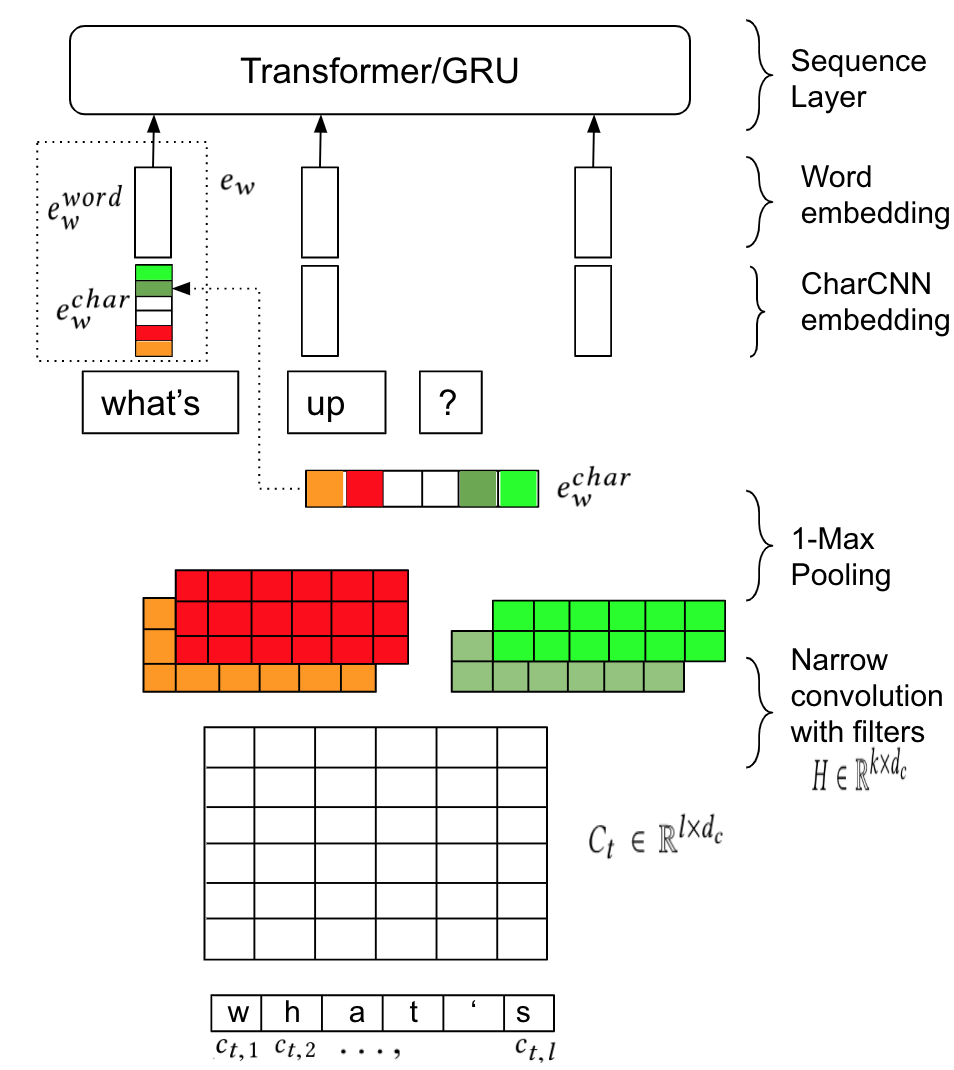}
    \caption{Illustration of encoder which comprises of a character based CNN layer for each word and GRU/Transformer layer at word level.}
    \label{fig:encoder}
\end{figure}

The architecture of the encoder is shown in Fig. \ref{fig:encoder}. Input to the encoder is a message $m_i$, consisting of a sequence of $N$ words $\{w_{i,1}, w_{i,2}, \ldots, w_{i,N}\}$, and the output is a dense vector $e_{i} \in \mathbb{R}^{d \times 1}$.

To represent a word, we use an embedding composed of 2 parts; a character based embedding aggregated from character representations, and a word level embedding to learn context representation. To generate the character based embedding, we use a character CNN \cite{kim2016character} that leverages sub-word information to learn similar representations for orthographic variants of the same word.
Let $d_c$ be the dimension of character embedding. For a word $w_{t}$ we have a sequence of $l$ characters $[ c_{t,1}, c_{t,2} \ldots, c_{t,l}]$. Then the character representation of $w_{t}$ can be obtained by stacking the character level embeddings in a matrix $C_{t} \in \mathbb{R}^{l \times d_c}$ and applying a narrow convolution with a filter $H \in \mathbb{R}^{k \times d_c}$ of $k$ width with \textit{ReLU} non linearity, and a max pooling layer. 
A $k$ width filter can be assumed to capture $k$-gram features of a word. We have multiple filters for particular a width $k$, that are concatenated to obtain a character level embedding $e_w^{char}$ for word $w_{t}$. The character level representation of the word is concatenated with the word-level embedding $e_w^{word}$ of $w_{t}$ to get a the final word representation $e_w$. Let $d_w$ be the dimenion of word embedding. So, the representation of a word $w_{t}$ becomes $ e_w \in \mathbb{R}^{(d_w + d_c) \times 1}$. 
\label{sec:encoder}
To capture the sequential properties of a message, we explore 2 architectures. \\
\textbf{Gated Recurrent Unit (GRU)} \cite{chung2014empirical} is an improved version of RNN to solve the vanishing gradient problem. 
We take the final step representation in the GRU to be the message embedding $e_i$.\\
\textbf{Transformer} \cite{vaswani2017attention} is an architecture solely based on attention to curb the recurrence step in RNN. It uses multi-headed attention to capture various relationships among the words. 
Similar to \citeauthor{yang2018learning} \shortcite{yang2018learning}, we use only the encoder part of transformer to create our message embeddings. 
We compute the message embedding $e_i$ by averaging the words vectors in the final layer of the transformer.
\subsection{Model Architecture}
\label{sec:model}
\begin{figure}[t!]
    \centering
    \includegraphics[width=6cm, height=4cm]{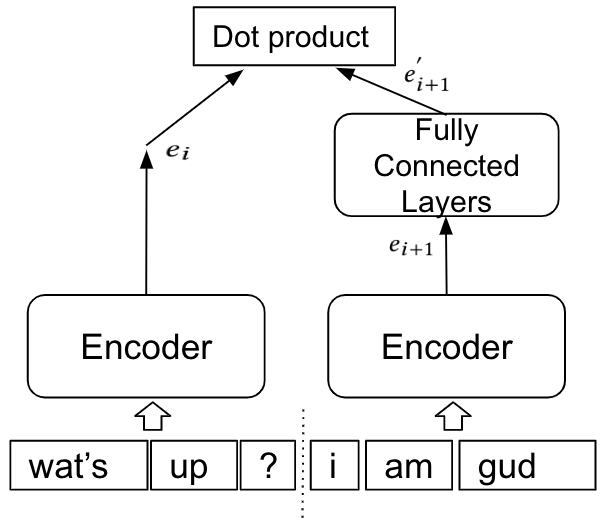}
    \caption{Model architecture for learning message embedding. 
    }
    \label{fig:M2}
\end{figure}

Akin to \citeauthor{yang2018learning} \shortcite{yang2018learning}, we use the model architecture shown in Fig. \ref{fig:M2} to train the encoders described in Sec. \ref{sec:encoder}. 
The input to our model is a tuple, $(m_i, m_{i+1})$, extracted from a conversation between 2 users. Messages $m_i$ and $m_{i+1}$ are encoded using the encoder described in Sec. \ref{sec:encoder} and represented as $e_{i}$ and $e_{i+1}$ respectively. Note that the parameters of both the encoders are tied, i.e they share weights. Hence, both $e_i$ and $e_{i+1}$ represent the encoding of message in same space. We transform embedding $e_{i+1}$ to reply space by applying 2 fully connected layers to obtain the response embedding $e_{i+1}^{'}$. Finally, the dot product of the input message embedding $e_{i}$ and the response embedding $e_{i+1}^{'}$ is used to score the replies. It maximizes the score of gold reply message higher in comparison to other replies. Within a batch, each $m_{i+1}$ serves as the correct response to its corresponding input $m_i$ and all other instances are assumed to be negative replies. It is computationally efficient to consider all other instances as negative during training because we don't have to explicitly encode specific negative examples for each instance.

\subsection{Message Clustering}
\label{sec:clustering}
We cluster frequent messages 
with the help of the embeddings learned as above. Our goal is to have a single cluster for all the orthographical variations and acronyms of a message.
We observe that the number of different variants of a phrase increases with the ubiquity of that phrase. For example, ``good morning'' has $\sim 300$ variants while relatively less frequent phrases tend to have significantly low number of variants. This poses a challenge when applying a standard density based clustering algorithm such as DBSCAN because it is difficult to decide a single threshold for drawing cluster boundaries. 
To handle this uncertainty, we choose HDBSCAN to cluster chat messages. This algorithm builds a hierarchy of clusters and handles the variable density of clusters in a time efficient manner. After building the hierarchy, it condenses the tree, extracts the useful clusters and assigns noise to the points that do not fit into any cluster. 
Further, it doesn't require parameters such as the number of clusters or the distance between pairs of points to be considered as a neighbour etc. 
The only necessary parameter is the minimum number of points required for a cluster. All the clusters including the noise points are taken to be the different classes in our message prediction step. 

\section{Message Cluster Prediction}
\label{groupprediction}
In this section, we describe our approach of predicting the message cluster that a user is going to send. 
We use previous received message in chat as the only context signal. Upon receiving a message, we predict a message that is a likely response. 
As the user starts typing, we update our message prediction and SR in real time after every character typed. The prediction latency should be in order of tens of milliseconds so that the user's typing experience is not adversely affected. 

We pose message prediction as a classification problem where we train a model to score the response messages. 
A neural network (NN) based classification model that accepts the last received message and the typed text as inputs can be a possible solution, but running inference on such a model on mobile devices proves to be a challenge \cite{gysel2016hardware}. The size of the NN model that needs to be shipped explodes since we have a large input vocabulary and a large number of output classes. The embedding layer that transforms the one-hot input to a dense vector usually has a size of the order of tens of MBs, which exceeds the memory limits that we set for our application. To overcome this issue, we evaluate 2 orthogonal approaches. One is to apply a quantization scheme to reduce the model size and the other is to build a hybrid model, composed of a NN component and a trie-based search. 

\subsection{Quantized Message Prediction Model}
\label{sec:quantization}
We train a NN for message prediction task where the input is the last received message and the typed text, the output is the message cluster scores. Top frequent $G$ message clusters that were prepared in Sec. \ref{sec:clustering} are chosen as classes for this model. We encode the last received message and typed text into a dense vector, which is fed as input to a classification layer that is a fully connected layer having sigmoid non-linearity and $G$ output neurons. Details of the training procedure is described in Sec. \ref{exp:prediction} 

An active area of research is to reduce the model size and the inference times of NN models with minimum accuracy loss so that they can run efficiently on mobile devices \cite{howard2017mobilenets,jacob2018quantization,ravi2017projectionnet}. One approach is to quantize the floating point representations of the weights and the activations of the NN from 32 bits to a lower number of bits. We use a quantization scheme detailed in \citeauthor{jacob2018quantization} \shortcite{jacob2018quantization} that converts both weights and activations to 8-bit integers and use a few 32-bit integers to represent biases. This reduces the model size by a factor of 4. When we quantize the weights of the network after training the model with full precision floats, the model accuracy reduce significantly. We employ quantize aware training\footnote{\url{https://github.com/tensorflow/tensorflow/blob/master/tensorflow/contrib/quantize/README.md}} to reduce the effect of quantization on model accuracy. 
Under this scheme, we use quantized weights and activation to compute the loss function of the network during training as well. This ensures parity during training and inference time. While performing back propagation, we use full precision float numbers because minor adjustments to the parameters are necessary to effectively train the model.
After qunatization, we obtain a model of size $\sim9$ MB . 

\subsection{Hybrid Message Prediction Model}
\label{sec:hybrid}
\begin{figure}[t!]
    \centering
    \includegraphics[width=\columnwidth]{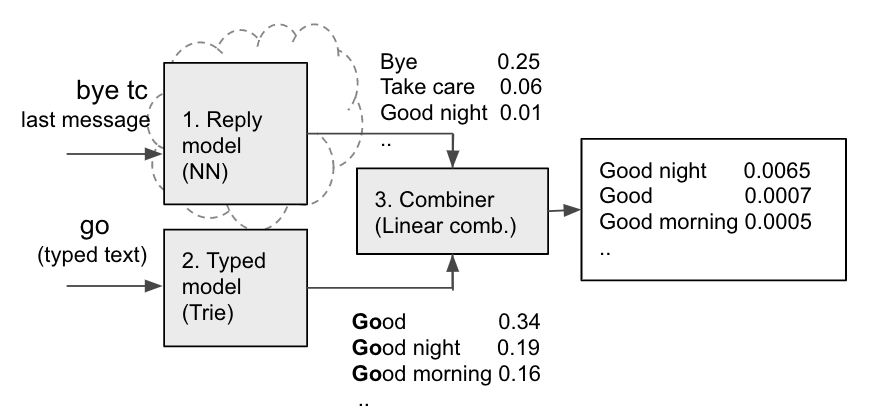}
    \caption{Example of hybrid model message cluster prediction. Message prediction from NN based reply model on server (1) is combined (3) with trie model prediction on client (2) for final message prediction score.}
    \label{fig:prediction}
\end{figure}

\begin{algorithm}[t!]
\caption{{\bf Message Cluster Prediction}}
\begin{flushleft}
\textbf{Input}: Last message received $prev$, Typed text $typ$
\end{flushleft}
\begin{algorithmic}[1]
\label{algo:pred}
\setlength{\abovedisplayskip}{1pt}
\setlength{\belowdisplayskip}{1pt}
\State Compute $P_{reply}(G= g| prev)$ with the reply NN model and send the message clusters scores to the client.
\State Using text $typ$ as input to Trie model, fetch set of phrases $S$ that start with $typ$. 
\State Compute trie score for message cluster $g$ as 
\begin{equation}\label{eq:trie}
\begin{split}
P_{trie}(g| typ) = \sum\limits_{ph \in S_g \cap S} \mathit{freq}(ph)/\sum\limits_{ph \in S}\mathit{freq}(ph)
\end{split}
\end{equation} where $S_g$ is the set of phrases in message cluster $g$.
\State Combine the scores for message cluster $g$ as:
\noindent
\hspace{-5mm}
\begin{equation}\label{eq:combiner}
\begin{split}
Q(g|prev, typ) & = (w_0 + w_{t}*P_{reply}(g| prev)) * \\
 & (w_{p_{0}} e^{-\lambda nc} + w_{p1} P_{trie}(g|typ) )
\end{split}
\end{equation}
\end{algorithmic}
\begin{flushleft}
\textbf{Output:} Message cluster scores $Q$ 
\end{flushleft}
\end{algorithm}

We build a hybrid message prediction model, where a resource intensive component is run on the server and its output is combined with a lightweight on-device model to obtain message predictions. 
Algorithm 1 outlines the score computation of the hybrid model and Fig. \ref{fig:prediction} demonstrates the high level flow with an example. The three components in our hybrid model are detailed below.\\
\textbf{Reply Model:}
Similar to the prediction model described in Sec. \ref{sec:quantization}, we build a NN model which takes the last received message $prev$ as input and outputs reply probabilities $P_{reply}(G=g | prev)$ for each message cluster $g \in G$. When a message gets routed to its recipient, the reply model is queried on the server and the response predictions are sent to the client along with the message. 
The message clusters that have a reply probability $P_{reply}(G=g|\text{prev})$ above a threshold $t_{reply}$ are sent to the clients.\\
\textbf{Typed Model:} 
Trie is an efficient data structure for prefix searches. We retrieve the relevant message clusters for a given typed text by querying a trie. It stores $<$phrase$>$ as key and $<$(message cluster id, frequency)$>$ tuple as value at leaf nodes. Phrase to message cluster id mapping is described in Sec. \ref{sec:groupformation}. 
We add the frequency of the phrases from our chat corpus as additional information in the trie, for scoring retrieved entries.
The score of a message cluster retrieved is calculated as shown in Eq. \ref{eq:trie}.
We create a trie with top frequent $34k$ phrases and $\sim 7500$ message clusters. In the serialized form, the size of the trie is around $700KB$. This is small enough for us to ship to client devices. Another advantage of trie is that it is interpretable and makes it easy for us to incorporate any new phrase, even if they are not observed in our historical data. \\
\textbf{Combiner:} 
A final score for a message cluster $ Q(g| \text{prev, typed}) $ is computed from $P_{reply}(g|\text{prev})$, $ P_{trie} (g| \text{typed})$ and length of string typed $nc$ as shown in Eq. \ref{eq:combiner}. The weighting of terms is designed such that as a user types more characters, contribution from the reply model vanishes unless the message cluster is not predicted from the trie. The weights are chosen by trial and error. 
\section{Message Cluster to Stickers mapping}
\label{groupmapping}
As mentioned in Sec. \ref{sec:intro}, we make use of a message cluster to sticker mapping to suggest suitable stickers from the predicted message clusters. When a sticker is created, it is tagged with conversational phrases that the sticker can possibly substitute.
We use this meta-data in order to map the message clusters to stickers. We compute the similarity between the tag phrases of a sticker and the phrases present in each message cluster, after converting them into vectors using the encoder mentioned in Sec. \ref{sec:groupformation}. Compared to the historical approach of suggesting a sticker when the user's typed input matches one of its tag phrases, the current system is able to suggest stickers even if different variations of the tag phrase are typed by the user, as we have many variations of a message already captured in the message clusters. We regularly refresh the message cluster to sticker mapping by taking into account the relevance feedback observed on our recommendations.
Since generation of message to sticker mapping is not the main focus of paper, we skip its details.
\section{Experiments}
\label{sec:experiment}

\frenchspacing
In this section, we first describe the dataset used for training the SR system. Next, we quantitatively evaluate different message embeddings on a manually curated dataset. 
We show qualitative results after clustering to show the effectiveness of the message embeddings. Then, we present a comparison of the NN model and the hybrid model for the message cluster prediction.
\subsection{Dataset and pre-processing} \label{sec:dataprepro}
To ensure user privacy in data collection, we strip user identity and replace it with anonymous ids. Further, we randomly sample $10\%$ of anonymous ids to collect the dataset for a period of $5$ months, from a particular geography for which we need to build a model. This will have a mixture of languages used in that region. This dataset is then pre-processed to extract useful conversations from the chat corpus. First step is tokenization, which includes accurate detection of emoticons and stickers, reducing more than 3 consecutive repetitions of the same character to 2. After pre-processing the data, we create tuples of the current and the next message for training the message embedding models. We get $~27m$ tuples after performing tokenization. Since stickers are mostly used to convey short messages, we filter out all the tuples which have at least one message having more than 5 words. After filtering, our input vocabulary consist of top frequent $50k$ words in our corpus. The dataset is randomly split into training and validation sets with $520k$ examples in validation set. 
We pad characters in words if the character length of word is less than $10$ characters. 

\subsection{Message Embedding Evaluation}
\label{sec:chat_abb_exp}
\begin{figure}[t!]
\includegraphics[height = 50mm]{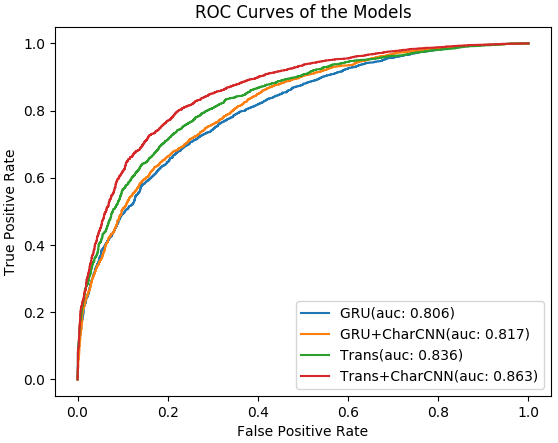} 
\caption{Performance of the message embedding models on phrase similarity task.}
\label{fig:graph1}
\end{figure}
In this subsection, we evaluate the embeddings generated from different architecture on a manually labeled data. First, we describe the baselines, training and hyper parameters.
\subsubsection{Message embeddings models:}
We compare the embeddings obtained from architecture (Sec. \ref{sec:model}) and its variants which are as follows: 1) {\em \bf Trans+CharCNN} - Encoder as shown in Fig. \ref{fig:encoder} with transformer. 2) {\em \bf Trans} - Uses only word-level embedding $e_m^{word}$ as input to transformer. 3) {\em \bf GRU+CharCNN} - Encoder as shown in Fig. \ref{fig:encoder} with GRU. 4) {\em \bf GRU} - Uses only word-level embedding as input to GRU.
\subsubsection{Training and Hyper-parameters:} We train our GRU based models using Adam Optimizer with learning rate of $.0001$ and step exponential decay for every $10k$ steps. We apply gradient clipping (with value of 5), dropout at GRU and fully connected layer to reduce overfitting. For the GRU+CharCNN model, we use filters of width $1,2,3,4$ with number of filters of $50,50,75,75$ for the convolution layers. The GRU models use embedding size of $300$.

Transformer models are trained using RMSprop optimizer with constant learning rate of $0.0001$. They use 2 attention layers and 8 attention heads. Within each attention layer, the feed forward network has input and output size of 256, and has a 512 unit inner-layer. To avoid overfitting we implement attention and embedding drop out of $0.1$. The convolution layers of the Trans+charCNN model are similar to those of GRU+CharCNN model. 

\subsubsection{Phrase Similarity Task:} We create a dataset which consists of phrase pairs labelled as similar or not-similar. For e.g. (``ghr me hi'', ``room me h''), (``majak kr rha hu", ``mjak kr rha hu") are labelled as similar while (``network nhi tha'', ``washroom gyi thi"), (``it's normal'', ``its me'') are labelled as not-similar. Possible similar examples are sampled from top 50k frequent phrases using 2 methods: a) Pairs close to each other in terms of minimum edit distance.
b) Pairs having words with similar word embeddings. 
Possible negative examples are sampled randomly. 
These pairs were annotated by 3 annotators. Pairs with disagreement within the annotators were dropped. Finally, the dataset has $3341$ similar pairs and $2437$ non-similar pairs.

To evaluate the models, we calculate the cosine similarity between the embedding of phrases in a pair. The ROC curves of the models are shown in Fig. \ref{fig:graph1}. We observe that transformer has significant improvement over GRU. This is due to better capturing of semantics of phrases using self-attention. CharCNN improves the performance of both, GRU and Transformer. It is able to capture chat characteristics which occur due to similar sounding sub-words. For e.g. `night' and `n8', `see' and `c' are used interchangeably during conversation. CharCNN learns such chat nuances and hence improves the performance. 
Trans+CharCNN model performs the best and shows $\approx$6\% absolute improvement in terms of AUC over the GRU baseline.

\subsection{Message Clustering Evaluation}
\label{sec:clustereval}

\begin{figure}[t!]
  \centering
  \includegraphics[width=\columnwidth]{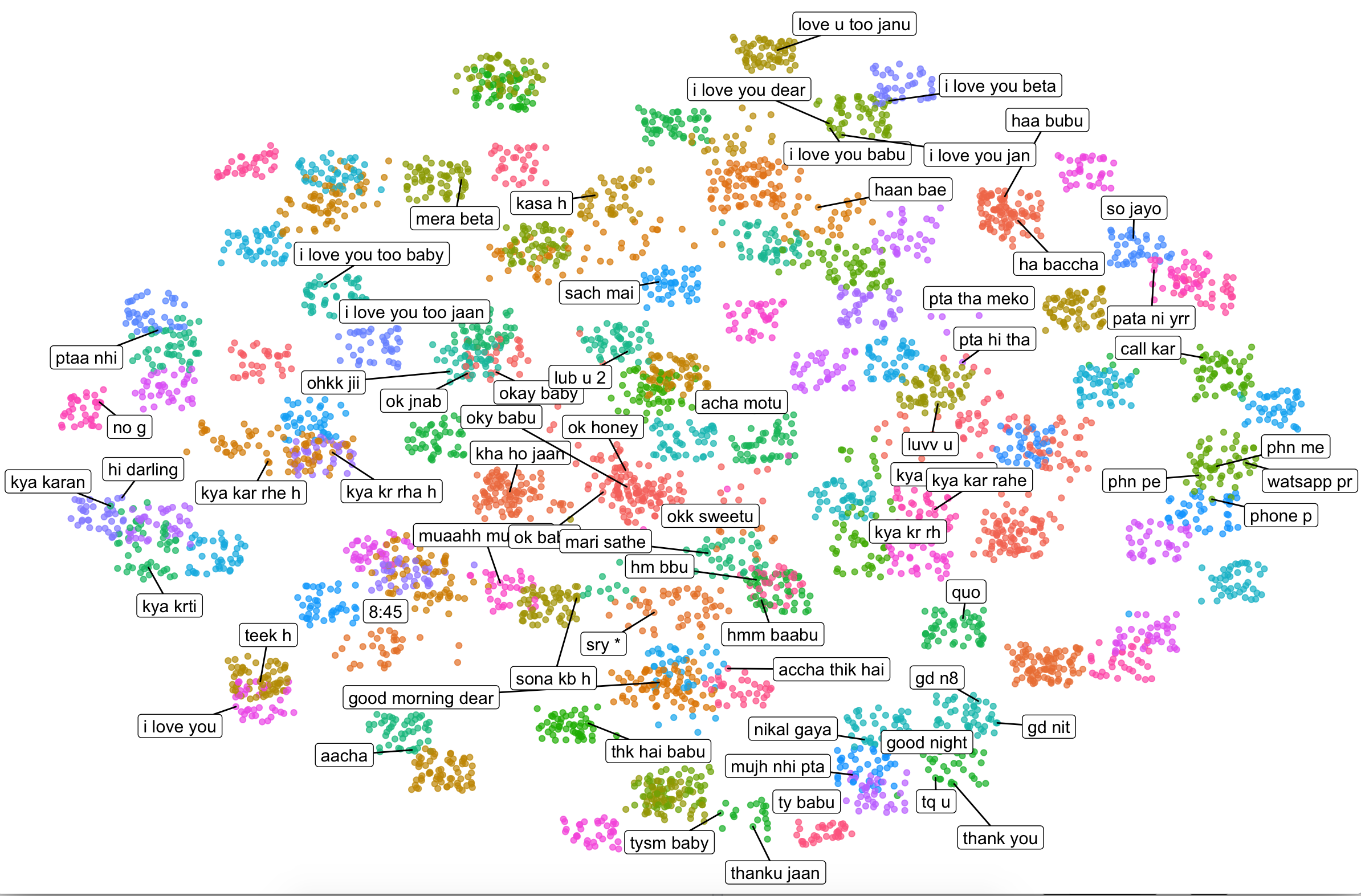}
\caption{HDBSCAN clusters of top frequent phrases using message embedding produced by GRU-CharCNN model. The points represents chat message vectors projected into 2D using TSNE. Read Section \ref{sec:clustereval} for more information. }
\label{fig:cluster_all}
\end{figure}

  

Fig. \ref{fig:cluster_all} shows a qualitative evaluation of clusters obtained from HDBSCAN algorithm. We select the top $100$ clusters based on the number of phrases present in cluster. We project the phrase embeddings learned from our model in 2-dimension using TSNE \cite{maaten2008visualizing} for visualization. Fig. \ref{fig:cluster_all} shows some randomly picked phrases from the clusters; phrases from the same cluster are represented with the same color. We observe that the various spelling variants and synonyms for a phrase are grouped in a single cluster. For example, the `good night' cluster includes phrases like `good nighy', `good nyt', `gud nite'. Our clustering algorithm is able to capture fine-grained semantics of phrases. For instance, `i love u', `i love you', `i luv u' phrases belong to one cluster while `love u too', `love u 2' belong to a different cluster which helps us in showing accurate SR for both clusters. If a user has messaged `i love u' then showing stickers related to `i love u too' cluster is more relevant replies as compared to showing `i love u' stickers. Our message embedding is able to cluster phrases like `i love u baby' and `i luv u shona' in a cluster distinct from the `i love u' cluster. This helps us deliver more precise SR since we have different stickers in our database for phrases like `i love u' and `i love u baby'.
\subsection{Message Cluster Prediction Evaluation}
\label{exp:prediction}
We compare our hybrid model and the quantized NN model on the following metrics. 1) Number of characters that a user needs to type for seeing the correct message cluster in top 3 positions (\# of Character to be typed), 2) How many times the model has shown a wrong message cluster before predicting the correct one in first 3 positions (\# of times inaccurate prediction) and 3) Fraction of messages that could be retrieved by a model in first 3 positions with a prefix of that message (Fraction of msg retrieved). Lower the number for the first 2 metrics, the better the model is; while the fraction of messages retrieved should be as high as possible.
\begin{table}[t]
\centering
\resizebox{\columnwidth}{!}{
\begin{tabular}{|c|c|c|c|}
\hline 
Method &\begin{tabular}[c]{@{}l@{}}\# of Character\\ to be Typed\end{tabular} & \begin{tabular}[c]{@{}l@{}}\# of times\\ inaccurate predictions \\               shown\end{tabular} & \begin{tabular}[c]{@{}l@{}}Fraction of msg\\ retrieved\end{tabular} \\ \hline
\begin{tabular}[c]{@{}l@{}}Quantized NN \\ (100d)\end{tabular} & \textbf{1.58} & 1.47 & 0.978 \\ \hline
Hybrid & 2.84 & \textbf{1.22} & \textbf{0.991} \\ \hline
\end{tabular}
}
\caption{\label{exp:msg-pred} Performance of the message prediction models}
\end{table}
For training the message prediction using NN model, we collect pairs of current and next messages from complete conversation data. Our training data had around $10M$ such pairs. We randomly sampled $38k$ pairs for testing. We curate the training data by treating all prefixes of the next message as typed text and its message cluster as the class label. We consider only top $7500$ message clusters on the basis of the total frequency of their phrases. Selected clusters cover $~34k$ top frequent reply messages in our dataset. The hybrid model is a combination of 2 models. In the first model, we predict the next message based on the current message. It is trained directly from pairs of consecutive messages, (current, next), in our corpus, where the next message was mapped to its corresponding message cluster. In the second model, we build a trie based on the typed message. We prepare phrase frequency of phrases as mentioned in Sec. \ref{sec:hybrid}.

Results of the various models are shown in Table \ref{exp:msg-pred}. The quantized NN model performs better in terms of \# of characters to be typed. This is expected because the NN model learns to use both inputs simultaneously whereas the combiner (last function in hybrid model) used is a simple linear combination of just three features. The hybrid model performs slightly better in terms of \# of times inaccurate recommendations shown and fraction of messages retrieved. Compared to the quantized model, the hybrid model needs approximately one more character to be typed on an average to get the required predictions. 
However, the quantized NN is $\sim 9.2$ MB in size, which goes beyond the permissible memory limits set in our use case. The on-device foot print for our hybrid model is below $1$ MB. Also, the trie based model can guarantee retrieval of all message clusters, with some prefix of the message. If the message is not interfered by another longer but more frequent message, the message class can be featured in top position itself, with some prefix of the message as input or with full message as input. A pure NN based model can't make such guarantees. Hence, more than $2\%$ improvement in the third metric shown in Table \ref{exp:msg-pred}. 
Given that this SR interface is one of the heavily used interfaces for sticker discovery, if a sticker is not retrieved through this recommendation, the user might think that the app doesn't support the message class or the app doesn't have such stickers. So, making the retrieval of message cluster close to $100\%$\% is critical for our SR models. 

\section{Deployed System}
There are multiple regional languages spoken in India. We segregate our data based on geography and train separate models (both message embedding and message prediction models) for each state. It ensures that frequent chat phrases distribution doesn't get skewed towards one majority spoken language across India. We obtain the primary language of a user from the language preferences explicitly set by them or infer it from their sticker usage.

We decompose our message prediction system in 2 steps. It helps in reducing the size of the model shipped to client. It also reduces the download failures on sketchy networks and data consumption. When a message is being sent from user A to user B, server fetches the response message cluster scores corresponding to the model assigned to user B. To maintain high speed of message delivery, we serve the reply model by caching the top $300k$ message in each geography corpus. For running typed model based on trie, we ship the asset files to client (trie and message cluster to sticker mapping). When a user logs in, client checks whether an updated model is present on server or not. If either trie or sticker mapping file has been updated, client downloads those files.

We observe that the top phrases in a corpus don't change much over time. Only certain event specific (\textit{e.g}. ``cricket world cup'', ``movies") or festival specific phrases get updated which mainly depend on seasonality. So, we freeze our message clusters for each geography and don't retrain the message embedding model. We only need to update frequency of phrases for message cluster prediction based on seasonality which we directly fetch from analytics. We don't require any model maintenance, since we are using unsupervised approach to generate clusters and only need the frequency of phrases for scoring the message cluster in trie. Complete pipeline for message prediction has been automated with anonymous chat corpus.
It helps us in extending to more languages and requires minimal manual efforts.


Using the method mentioned in section \ref{groupmapping}, we are able to add a new stickers to the system with the help of very few tags which would have been already decided at the time of its creation. The sticker mappings thus created are later scored and ranked with the help of sticker usage data across users.

 \noindent \textbf{User Impact:} The proposed system has been deployed in production on Hike for more than 6 months. Before a full rollout, we conducted state wise A/B experiment because models are trained separately for each state.
 We chose user sets of size, of the order of tens of thousands as control and test in each experiment group.
A/B tests to compare this system with a previous implementation showed an average $\sim8\%$ relative improvement in the fraction of users who send stickers, among those who send a message. Compared to legacy system, proposed SR system also increases the volume of stickers exchanged.

\section{Conclusion \& Future Work}\label{sec:conclusion}

\frenchspacing
In this paper, we present our deployed system for deriving contextual, type-ahead SR within a chat application. We decompose the task into two steps. First, we predict the next message that a user is likely to send based on the last message and the typed text. Second, we substitute the predicted message with relevant stickers. 
We discuss how numerous orthographic variations for the same utterance exist in Hike's chat corpus, which mostly contains messages in a transliterated form. 
We describe a clustering solution that can identify such variants with the help of message embedding, which learns similar representations for semantically similar messages.
Message clustering reduces the complexity of the classifier used in message prediction. \textit{E.g.}, by predicting one of 7500 classes (message clusters), it is able to cover all intents expressed in $\approx$34k frequent messages. 
For message prediction on low-end mobile phones, we propose a hybrid model that combines a NN on the server and a memory efficient trie search on the device for low latency SR. We show experimentally that the hybrid model is able to predict a higher fraction of overall messages clusters compared to a quantized NN. This model also helps in better sticker discovery for rare message clusters. Our described system has been deployed for $8$ Indian languages and serving millions of users daily with $\sim8\%$ relative increase in the fraction of users sending stickers. 

In the future, we plan to add character level convolution network for message prediction on client which is memory efficient compare to current quantized NN. Sticker mapping can also be improved by adding other signals like user's sticker preference \textit{etc}. Though we construct the message clusters to reduce complexity of the message prediction task, we observe that these message clusters are generic enough to be used in other application such as conversational response generation, intent classification \textit{etc.} \cite{serban2016building}.
\bibliographystyle{aaai}
\fontsize{9pt}{10pt}\selectfont
\bibliography{iaai2020}
\end{document}